# Machine Learning-based Prediction of Porosity for Concrete Containing Supplementary Cementitious Materials

Chong CAO*

University of California, Los Angeles, 110 Westwood Plaza, Los Angeles, CA 90095

*Corresponding author at:

UCLA Anderson School of Management, 110 Westwood Plaza, Los Angeles, CA 90095

Tel: +1 (310) 779 7537

Email address: jasongo@ucla.edu

# Abstract

Porosity has been identified as the key indicator of the durability properties of concrete exposed to aggressive environments. This paper applies ensemble learning to predict porosity of high-performance concrete containing supplementary cementitious materials. The concrete samples utilized in this study are characterized by eight composition features including *w/b* ratio, binder content, fly ash, GGBS, superplasticizer, coarse/fine aggregate ratio, curing condition and curing days. The assembled database consists of 240 data records, featuring 74 unique concrete mixture designs. The proposed machine learning algorithms are trained on 180 observations (75%) chosen randomly from the data set and then tested on the remaining 60 observations (25%). The numerical experiments suggest that the regression tree ensembles can accurately predict the porosity of concrete from its mixture compositions. Gradient boosting trees generally outperforms random forests in terms of prediction accuracy. For random forests, the *out-of-bag* error based hyperparameter tuning strategy is found to be much more efficient than *k*-Fold Cross-Validation.

***Keywords***: Machine learning; Decision trees; Ensemble learning; Concrete durability; Porosity

# 1 Introduction

Concrete is a composite material made of aggregates and hydrated cement paste that each may contribute to the formation of interconnected pore structure. There is considerable interest in the relationship between porosity characteristics and transport properties of concrete, such as diffusion coefficient of oxygen and carbon dioxide, gas/water permeability, chloride ions migration, and electrical resistivity [1-11]. The empirical relationship between the compressive strength and the transport properties of concrete (expressed as intrinsic permeability) as a function of the capillary porosity is shown in Fig. 1. Generally, the increase in porosity will increase the permeability of concrete and reduce the mechanical strength. Therefore, porosity has been identified as one of the key parameters in determining the durability and serviceability of concrete structures subjected to aggressive environments [12].

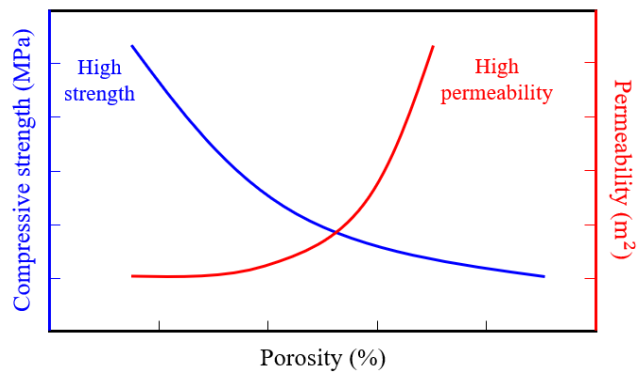

**Fig. 1**. Concrete porosity and durability properties

Among the many factors that may affect the porosity of concrete, water/cement ratio ($w/c$) plays an essential role in facilitating the hydration reactions of cement paste. The volume of the capillary pores in the hydrated cement paste increases with the $w/c$ ratio. As the curing days increase and the hydration proceeds, the porosity decreases as a result of the reduction in large-dimension pores that have been filled or connected by calcium-silicate-hydrate (C-S-H) gel pores. Powers proposed the classical model to calculate the volumetric composition of hardened cement paste from *w/c* ratio and degree of hydration of the cement [13]. Concrete can become more heterogeneous due to the presence of aggregates and the interfacial transition zone (ITZ). Previous experimental results show that the increase in size and proportion of coarse aggregate will lead to the increase in the local porosity at the ITZ and thus the reduction in the overall durability properties of ordinary concrete [14]. The ratio of coarse aggregate to fine aggregate by weight (CA/FA) is found to be a critical factor influencing porosity, tortuosity and permeability of concrete [15].

Supplementary cementitious materials (SCMs) have been utilized to partially replace Portland cement for the purpose of enhancing durability and strength properties of concrete [16-19]. The use of SCMs such as fly ash, a byproduct of the combustion of coal powder in thermoelectric power plants, or ground granulated blast-furnace slag (GGBS), a byproduct of pig iron production, may also promote cleaner production by significantly reducing $CO_2$ emission. The beneficial effect of GGBS in concrete lies in the latent hydraulic reaction that contributes to the cement hydration process by densifying the concrete matrix and refining the pore structure. This will lead to reduction in porosity and increase in compressive strength of concrete at later ages. The change in mineralogy of the cement hydrates may also improve the chloride binding capabilities and increase the electrical resistivity of concrete [20]. The reduction in the permeability of fly ash concrete has been attributed to a combination of the reduced water content for a given workability and the refinement of pore structure due to pozzolanic reaction. Because of the long-term nature of the pozzolanic reaction, the beneficial effects associated with it become more evident in well-cured concrete, and therefore the curing condition (air curing or water curing) will be another crucial factor influencing porosity of high-performance concrete [21]. Furthermore, the addition of superplasticizers (SP) may allow for substantial reduction in the mixing water, which will facilitate the formation of denser pore structure [22].

Due to the various mixtures of concrete and the time-dependent hydration process of cement, the development of pore structure within concrete becomes very complex, which may defy analytical modeling. The major difficulty lies in the uncertainties associated with the pozzolanic and hydraulic reactivity of fly ash and slag. The chemical composition of SCMs can vary significantly and the estimation of the reactive portion of the materials is quite challenging. Papadakis [23,24] has proposed a theoretical model to predict the chemical and volumetric composition of fly ash concrete. The model considers the stoichiometry of Portland cement hydration and pozzolanic reactions of fly ash as well as the molar weights of reactants and products. However, this model assumes full hydration of Portland cement and the complete pozzolanic reactions of fly ash and therefore can't consider the time-dependent evolution of porosity. This motivates the data-driven approach to be widely adopted in modeling high-performance concrete properties.

In order to make empirical prediction of the permeation properties of high-performance concrete, Khan [25] applied multivariate regression to predict porosity based on concrete mixtures such as fly ash proportion, microsilica content, and water/binder ratio ($w/b$) at different ages of 28, 90 and 180 days. An alternative statistical approach for modelling concrete with complex mixture compositions would be machine learning, which has been widely used to predict the mechanical and durability properties of high-performance concrete with highly desirable accuracy [26-31]. However, little literature deals with the machine learning-based prediction of concrete porosity. Boukhatem *et al*. [32] presented a Neural Network modeling framework to predict compressive strength, porosity and transport tortuosity of fly ash concrete using mix design parameters (water, binder, aggregates, superplasticizer), fly ash content and age as input. The modeling results showed excellent correlation between predicted values and experimentally obtained porosity, which suggests that machine learning is a promising technique for predicting concrete porosity.

Among all the well-known machine learning methods, decision trees have emerged as the most popular supervised learning approach for data mining. Decision trees can naturally incorporate mixtures of numeric and categorical predictor variables and missing values. They are insensitive to the monotone transformation of the individual predictors and immune to the effects of predictor outliers. Moreover, decision trees are able to handle many irrelevant inputs because they perform internal feature selection as an integral part of the algorithm. Furthermore, ensemble learning can dramatically improve the prediction accuracy of decision trees by aggregating multiple weak learners [33]. Therefore, with their robust predictive performance and high interpretability, ensemble trees have been widely used to characterize concrete properties [34-38].

The objective of this paper is to apply regression tree ensembles for empirically predicting the porosity of high-performance concrete containing SCMs. First, a reliable database consisting of 240 data records for concrete porosity is assembled from published literature. The full dataset is randomly divided into 75% training dataset and 25% testing dataset. Then, both gradient boosting trees and random forests are trained to tune key hyperparameters by minimizing either $k$-Fold Cross-Validation error or *out-of-bag* error through Bayesian optimization algorithm. Finally, the optimized models will be tested on the testing data set in order to gain a measure of the prediction accuracy. Special attention will be given to the estimation of predictor importance based on ensemble trees. The proposed case study is utilized to compare the data-driven approach with classical chemo-mechanical model for the prediction of concrete porosity.

# 2 Experimental Database

In order to establish a reliable database for the development of the machine learning-based prediction model, experimental data of concrete porosity have been collected from published literature with certain selection criterion. First, the concrete is mixed with ordinary Portland cement (OPC), which can be partially replaced by either fly ash or GGBS. Second, all concrete specimens contain coarse aggregate and fine aggregate. Third, the curing regime falls into two categories: air curing and water curing. Fourth, the effect of carbonation on pore structure refinement is not introduced in the concrete specimens. Also, it has been determined to retain a balanced proportion for each type of concrete, i.e. ordinary Portland cement concrete (OPC), fly ash concrete and GGBS concrete.

The assembled dataset consists of 240 data records, featuring 74 unique concrete mixture designs [39-44]. Selected experimental data of concrete composition and porosity are shown in Table 1. There are eight input features: $w/b$ ratio, binder content (kg/m$^3$), fly ash content (%), GGBS content (%), superplasticizer content (%), CA/FA ratio, curing condition (categorical predictor) and curing days. The fly ash and GGBS contents are recorded as replacement fraction of the binder (cement plus SCMs). To ensure the consistency of the database, the applied dosage of superplasticizers is reported as weight proportion of the binder used in the concrete mixture. If the SP content is recorded as volume values in the original literature, a density of 1.2 kg/L is assumed for SP in order to translate volume numbers into weight proportion. There are no missing values for all 8 predictors in the established database. The statistics pertaining to each of the continuous variables are summarized in Table 2.

Curing days and *w/b* ratio are generally considered as the critical factors influencing concrete porosity. The visualization of porosity data against the two key parameters for different types of concrete (75 data records for OPC, 45 data for GGBS and 120 data for Fly ash) is shown in Fig. 2. There is no clear trend observed for both parameters in relation to porosity. This highlights the difficulty of predicting the porosity properties of high-performance concrete.

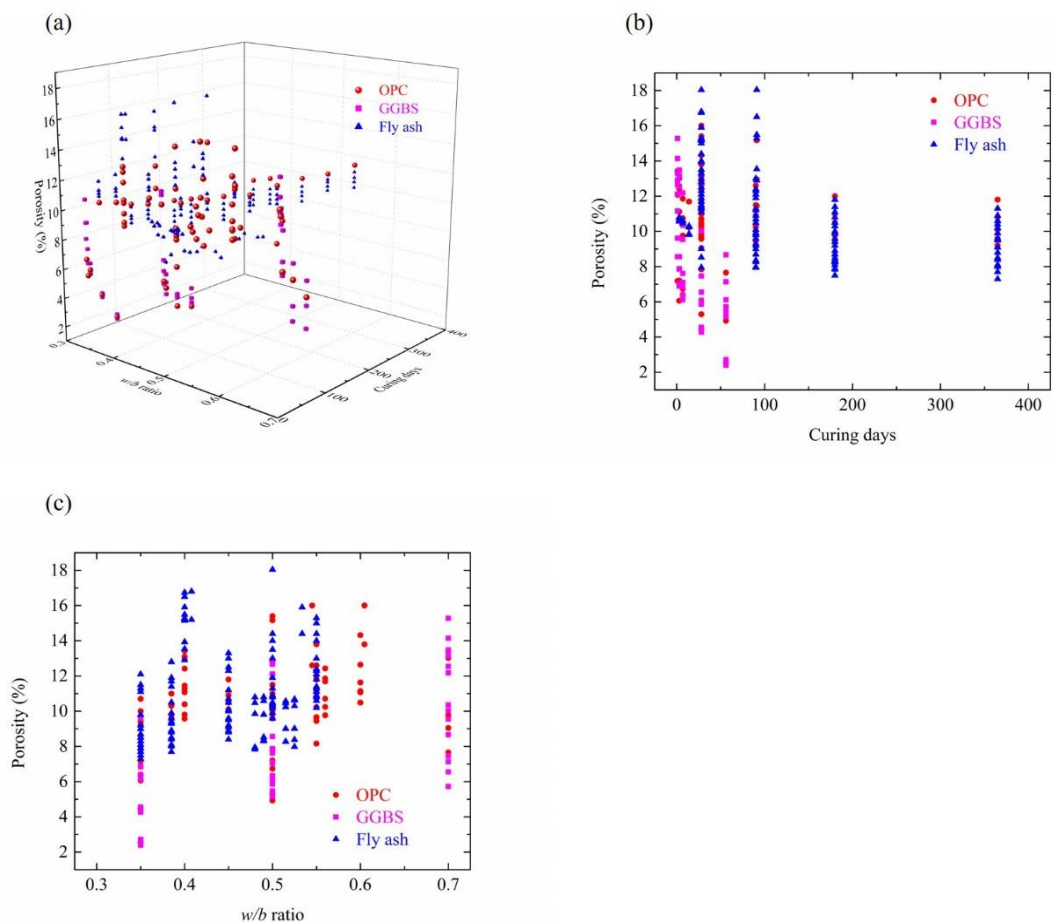

**Fig. 2**. Concrete porosity data: (a) 3D visualization; (b) Effect of curing days; (c) Effect of *w/b* ratio

In the original experimental work, the concrete porosity was measured using three distinct methods: water saturation under vacuum, mercury intrusion porosimetry (MIP) and helium porosimeter. These approaches, however, are based on different theoretical background and therefore may not give the same porosity interpretations. A further discussion of the possible discrepancy resulting from different measuring techniques is beyond the scope of this study.

For the purpose of testing the prediction accuracy of the machine learning algorithm, the full dataset is randomly divided into two groups: training dataset (75%) and testing dataset (25%). An identifier column is created (column name "Training" in Table 1) to indicate which subset the data instances belong to. Stratified sampling has been employed to reduce the sampling bias associated with the random partition of the training dataset and the holdout testing dataset. The training dataset should contain the various categories of concrete in similar proportions to the overall dataset. For each concrete type, the distribution of curing days in the training dataset should be representative of that in the whole dataset. Also, it is necessary to make sure that the input variables (e.g., *w/b* ratio and binder content) of the training group contain values that span the entire range of the overall dataset.

The categorical predictor "curing condition" is not transformed into numeric values, because decision trees can directly handle combinations of numeric and categorical predictor variables. The standardization of datasets is commonly required for many machine learning algorithms if the values of predictors vary on significantly different scales. Note that decision trees (including random forests and gradient boosting) are not sensitive to the magnitude of variables. Therefore, standardization is not needed before fitting ensemble trees.

# 3 Machine Learning Method

## *3.1 Regression tree ensembles*

Regression trees is a top-down, greedy approach that performs recursive binary splitting to grow a large tree on the training data set, stopping only when the terminal node has reached certain minimum number of observations. At each split, the best partition of predictor space is found by minimizing the sum of squared residuals (RSS) for the resulting predictions. The optimum number of terminal nodes can be obtained by applying cost complexity pruning to the large tree for a trade-off between the subtree's variance and bias. For a given test observation, the prediction can be made by using the mean response for all the training observations within the terminal node to which that test observation belongs [45].

Bootstrap aggregating (Bagging) is an important ensemble learning technique to reduce the variance of decision trees [46]. To apply bagging to regression trees, many bootstrapped replicas of the original training data set are first generated by repeatedly random sampling with replacement. Then, separate predictions are made by constructing $B$ independent regression trees using $B$ different bootstrapped training data sets. The final prediction can be obtained by averaging all the resulting predictions as follows

$$\hat{f}_{\text{bag}}(x) = \frac{1}{B}\sum_{b=1}^{B}\hat{f}^{*b}(x) \tag{1}$$

where $\hat{f}^{*b}(x)$ represents the prediction obtained from the $b^{\text{th}}$ bootstrapped training data set, and $B$ is the total number of bootstrapped training data sets. Using a sufficiently large number of bagged trees ($B$) can significantly reduce the prediction error without worrying about overfitting.

In addition, every tree in the ensemble can randomly select a subset of predictors for each decision split, a technique called Random Forests known to improve the accuracy of bagged trees when applied to high-dimensional data set consisting of highly correlated predictors [47]. As per the inventors' recommendations, the default value for the number of predictors to select at random for each split is approximately equal to one third of the total number of predictors for regression (minimum node size of 5) or square root of the total number of predictors for classification (minimum node size of 1) [33]. In practice, the best value for the number of predictors to select will depend on the problem and therefore should be treated as a tuning parameter. The simple example of bagging two regression trees in Random Forests is visualized in Fig. 3.

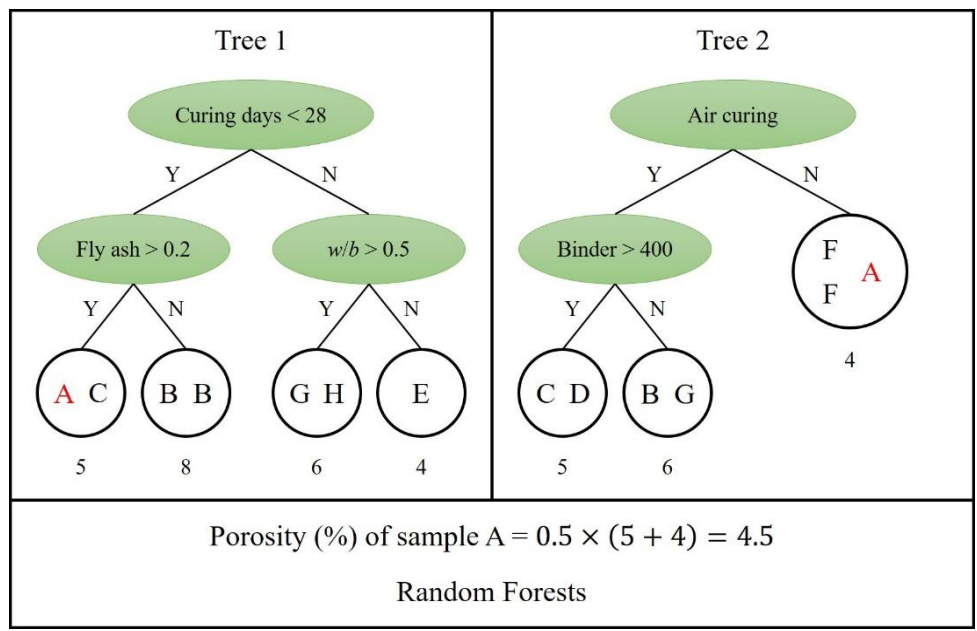

**Fig. 3**. Bagging regression trees

Boosting is another popular approach for improving the prediction accuracy of decision trees. Unlike bagging, gradient boosting does not involve bootstrap sampling or growing independent regression trees, but constructs new tree to the residual errors resulting from the aggregated prediction of all trees grown previously. The new decision trees, usually small with just a few terminal nodes, are sequentially added into the fitted function in order to update the residuals. This vanilla gradient boosting algorithm is often referred to as Least-squares Boosting (LSBoost), which applies gradient boosting on squared-error loss function so that the negative gradient is just the ordinary residual [48]. The LSBoost algorithm for regression is briefly illustrated in Fig. 4. The final output of the boosted regression tree ensembles can be obtained as the weighted sum of the predictions from all trees

$$\hat{f}_{boost}(x) = \sum_{b=1}^{B} \lambda \hat{f}^{b}(x) \quad (2)$$

where $\hat{f}^{b}(x)$ represents the prediction resulting from the $b^{th}$ residual correcting tree, $B$ is the total number of boosted trees, and $\lambda$ denotes the shrinkage parameter that controls the residual updating process. Unlike Bagging and Random Forests, Boosting can overfit if the total number of boosted trees ($B$) is too large. In the case of overfitting for gradient boosting, typical remedial measures may include reducing the number of learning cycles, limiting the tree depth and decreasing the learning rate ($\lambda$).

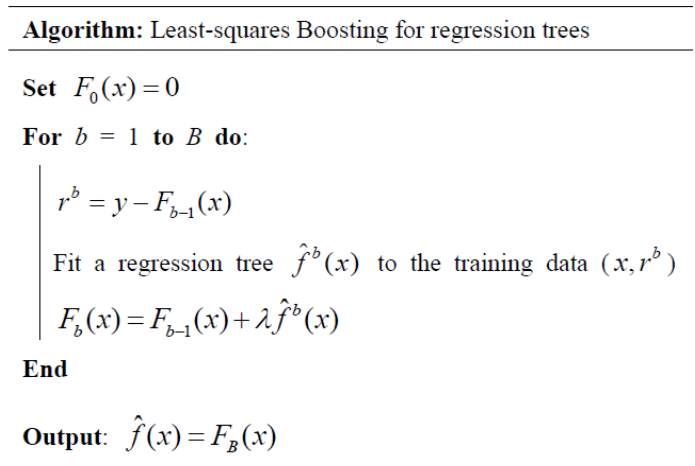

**Fig. 4**. Least-squares Boosting algorithm for regression trees

## 3.2 Hyperparameter tuning

In order to achieve best prediction performance for the machine learning process, hyperparameters should be tuned by optimization algorithms, such as grid search, randomized search and Bayesian optimization. The hyperparameters for ensemble trees include number of trees, maximum number of decision splits per tree, minimum number of terminal node observations, learning rate for shrinkage (for LSBoost), and number of predictors to select at random for each split (for Random Forests). The typical hyperparameters for both Random Forests and Gradient Boosting and their tuning range considered in the present study are listed in Table 3. Note that the values considered for each integer variable include the entire grid within the tuning range.

Learning rate ($\lambda$) is a critical regularization parameter for gradient boosting to shrink the contribution of each individual tree grown in the ensemble. Small learning rate (more shrinkage) will typically require a large number of learning cycles in order to achieve good prediction performance. However, overfitting may occur if the number of trees ($B$) is too large. The computation time increases linearly with the number of learning cycles. In this study, the number of trees is set to be tunable in the range of [10, 500] for LSBoosting and the learning rate can be chosen between 0.001 and 1. In the case of Random Forests, the number of trees is typically not tunable but should be set large enough for convergence. The prediction performance measures, e.g. mean squared error, generally show monotonously decreasing pattern with the increasing number of trees [49]. Moreover, the more trees are trained, the more stable the predictions should be for the variable importance [50]. Thus, a sufficiently large number of trees ($B = 300$) is chosen for the regression forest.

The maximum number of decision splits per tree is often referred to as the maximum tree depth, which determines the complexity of the individual trees. The deeper a tree, the more likely it overfits the training data set. But a too shallow tree might not allow for enough feature interactions. Another factor that controls the tree depth is the minimum number of terminal node observations. The smaller the minimum terminal node size, the deeper the trees can grow. The default values of the tree depth controllers for boosting regression trees are 10 for maximum number of splits per tree and 5 for minimum terminal node size in MATLAB [51], which indicate that shallow decision trees are grown. In this study, the maximum number of splits per tree is limited to be in the range of [1, 20] for LSBoosting. Meanwhile, the minimum size of terminal node observations is set to be tunable between 1 and 90 by default.

Random Forests, on the other hand, can achieve good prediction performance by simply using full-grown trees, according to its inventor' arguments [33]. The default maximum number of decision splits is $n$-1 for bagging tree ensembles in MATLAB, where $n$ is the number of observations in the training data set. The minimum number of observations per tree leaf is set as 5 by default. The experimental study by Segal [52] suggests that prediction performance gains can be realized by controlling the depth of the individual trees grown in the regression forest, especially for certain data sets with large number of noise variables. In this study, to find the optimal tree depth for Random Forests, the minimum terminal node size is treated as a tunable hyperparameter and specified to be at most 20. The maximum number of splits per tree is using the default parameter, which is 179 for the training data set of concrete porosity.

The number of predictors to select at random for each split is a critical parameter for Random Forests. Smaller number of randomly drawn candidate variables leads to less correlated trees, which may yield better stability when aggregating. This works particularly well when there are a large number of correlated predictors. However, for high dimensional data with only a small fraction of relevant variables (e.g. genetic datasets [53]), Random Forests is likely to perform poorly with a small subset of predictors. This is because the subtrees with selected groups of irrelevant variables may add additional noise into the trees and therefore reduce the ensemble prediction accuracy. For the concrete porosity data set considered in this study, the number of predictors to select at random for each split is set to be tunable in the range of [1, 8]. For boosted tree, all predictors should be selected at each split in order to precisely analyze the predictor importance [51].

## *3.3 Optimization algorithm*

To optimize the hyperparameters for the machine learning algorithm, *k*-Fold Cross-Validation (CV) is commonly employed to train the dataset and estimate the prediction error. This approach involves randomly dividing the entire training data set into *k* distinct groups (folds) of approximately equal size. With *k*-1 folds of the observations treated as the training data set, the mean squared error (MSE) of prediction can be computed on the observations in the hold-out fold (testing data set). After repeating the procedure for *k* times, the *k*-fold CV loss estimate is computed by averaging these values

$$\mathrm{CV}_{(k)} = \frac{1}{k}\sum_{i=1}^{k}\mathrm{MSE}_i \qquad (3)$$

where *k* is typically chosen as 5 or 10 for the bias-variance trade-off [45]. The hyperparameters for ensemble trees can be optimized by minimizing the *k*-fold CV loss. By default, the optimization objective function for regression is log(1 + 10-fold CV loss) in MATLAB [51]. In the present study, the gradient boosted trees (LSBoost) are optimized through 10-fold CV.

An important feature of bagging is that it offers a computationally efficient way to estimate the test error, without the need to perform cross-validation. For each of the bagging iterations, approximately 63.2% of the original training data set is selected as the bootstrapped sample [33].

$$\Pr\{\text{observation } i \in \text{bootstrap sample } b\} = 1 - \left(1 - \frac{1}{n}\right)^n \approx 1 - e^{-1} = 0.632 \qquad (4)$$

The remaining one-third of the observations that are not used to fit a given bagged tree, aka *out-of-bag* (OOB) observations, can be used as a test set to get a measure of the prediction error [54]. First, by running the entire bagging cycles, roughly $B/3$ predictions (on average) can be made for the $i$th observation using each of the trees in which that observation is OOB. Then, for each of the $n$ observations in the training data set, the corresponding predicted responses can be averaged to obtain the OOB predictions. Finally, the overall OOB mean squared error can be conveniently computed as [33]

$$\text{OOB MSE} = \frac{1}{n}\sum_{i=1}^{n}\frac{1}{|C^{-i}|}\sum_{b \in C^{-i}}\left[y_i - \hat{f}^{*b}(x_i)\right]^2 \qquad (5)$$

where $C^{-i}$ is the subset of indices of the bootstrapped sample $b$ that do not contain observation $i$, and $|C^{-i}|$ is the total number of such samples. To ensure that $|C^{-i}|$ is greater than zero, only observations that are out-of-bag for at least one tree are considered.

In this study, the hyperparameters for a random forest of regression trees (i.e., minimum size of terminal nodes and number of features to select at each node) are tuned by minimizing the OOB mean squared error. A custom objective function that accepts the tuning parameters as inputs is defined in MATLAB to compute the ensemble OOB MSE on the training data set. Similar hyperparameter tuning strategy based on *out-of-bag* predictions has been employed by Probst *et al.* [50] to achieve faster computing.

The Bayesian optimization algorithm has been adopted in this study to search for the best combination of hyperparameters for regression tree ensembles. Bayesian optimization typically works by assuming Gaussian process (surrogate model) for the objective function and maintains a posterior distribution for this function as the results of running machine learning algorithm experiments with different hyperparameters are observed [55]. One unique feature of Bayesian optimization is the acquisition function, which the algorithm uses to determine the point to evaluate in the next iteration. The acquisition function estimates the expected amount of improvement in the objective function over the currently available best result. It can also balance the tradeoff between exploration of new instances in the areas that have not yet been sampled and exploitation of the already examined area based on the current posterior distribution.

The basic procedures for Bayesian optimization are summarized as follows:

(a) Start with initial point of hyperparameter setting to evaluate the objective function by running machine learning algorithm experiment.

(b) Update the Gaussian process (surrogate model) to obtain a posterior distribution over the target objective function.

(c) Pick the next point of hyperparameter setting for evaluation by maximizing the acquisition function of expected improvement over the current best result.

(d) The procedure is repeated and the algorithm stops after a certain number of iterations (default 30 in MATLAB).

Bayesian optimization offers a natural framework for model-based global optimization of noisy, expensive black-box machine learning algorithms [56]. Compared with grid search and randomized search, Bayesian optimization is considerably more efficient as it can detect the optimal hyperparameter combinations by analyzing the previously-tested values, and running surrogate model is much cheaper than optimizing the objective function.

## 3.4 Prediction performance evaluation

Once the optimum hyperparameters have been obtained for the machine learning models, the prediction performance can be evaluated on the hold-out testing dataset. To compare the performance of gradient boosted trees and random forest, three statistical parameters are used to measure the prediction accuracy.

The root mean squared error (RMSE) for prediction is calculated as

$$\text{RMSE} = \sqrt{\frac{1}{m}\sum_{i=1}^{m}\left[y_i - \hat{f}(x_i)\right]^2} \tag{6}$$

where $m$ is the number of observations in the testing data set and $\hat{f}(x_i)$ gives the prediction for the $i$th observation.

The mean absolute percentage error (MAPE) is given by the following equation

$$\text{MAPE} = \frac{1}{m}\sum_{i=1}^{m}\left|\frac{y_i - \hat{f}(x_i)}{y_i}\right| \times 100\% \tag{7}$$

The $R^2$ statistic measures the proportion of variability in the response that can be explained by performing the regression

$$R^2 = 1 - \frac{\sum_{i=1}^{m}\left[y_i - \hat{f}(x_i)\right]^2}{\sum_{i=1}^{m}\left[y_i - \bar{y}\right]^2} \qquad (8)$$

Generally, an $R^2$ statistic that is close to 1 indicates good performance for the regression.

# 4 Results and Discussion

## 4.1 Experimental results

For Random Forest, the observed concrete porosities are compared with predicted values for both training and testing data sets in Fig. 5a and b. The optimal minimum size of terminal nodes is 1, which indicates that full-depth trees have been grown for Random Forest without the worry of overfitting. On the other hand, the best number of features to select at random for each split is 8, which means that all predictors should be selected at each split and Random Forest is reduced to Bagging in this case.

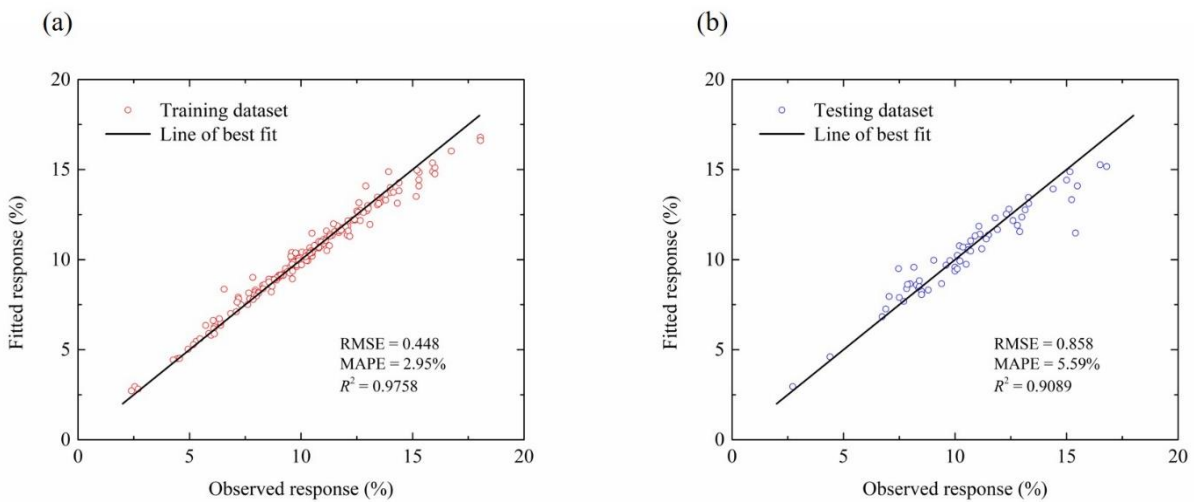

**Fig. 5**. Prediction performance of random forests: (a) Training dataset; (b) Testing dataset

To confirm that the number of trees grown in the regression forest is sufficient for achieving optimal prediction accuracy, the *out-of-bag* error is plotted against the number of trees in Fig. 6. It can be clearly observed that the OOB MSE is monotonically decreasing with number of trees and has settled down after growing 100 trees.

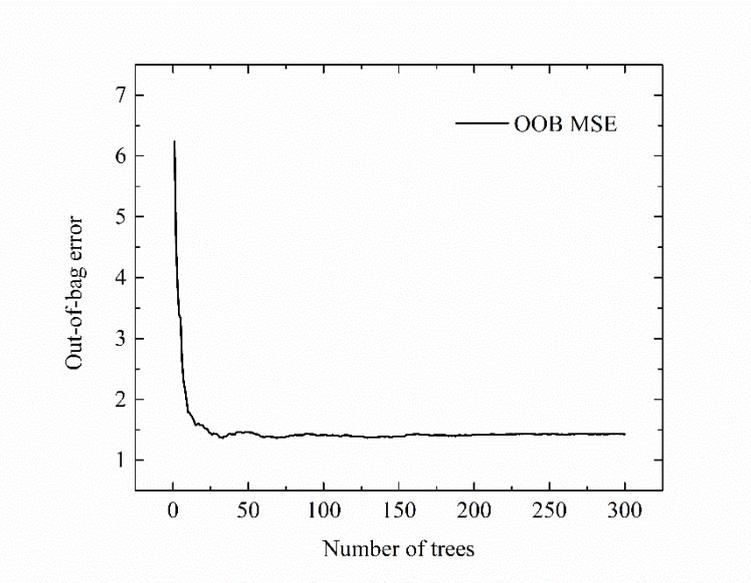

**Fig. 6**. *Out-of-bag* (OOB) error plot for random forests

In the case of gradient boosted trees, the observed concrete porosities are compared with predicted values for both training and testing data sets in Fig. 7a and b. The best learning rate is obtained as 0.1404 and the corresponding number of learning cycles is 486. The optimal maximum number of splits is 7, which confirms that shallow trees are grown in the ensemble. This highlights one difference between boosting and bagging: in boosting, because the growth of a particular tree takes into account the other trees that have already been grown, smaller trees are typically sufficient. The optimal minimum number of observations on terminal nodes is 5.

For the purpose of comparing random forest and gradient boosting on the concrete porosity data set, the prediction performance statistics based on 100 simulations are summarized in Table 4. It can be clearly seen that gradient boosting outperforms random forest in terms of average and best prediction accuracy (RMSE, MAPE and $R^2$ statistics). Similar observations have been reported by Chou *et al*. [28], who have compared the performance of different data-mining techniques for predicting concrete compressive strength from mixture compositions. The experimental results show that the optimized Multiple Additive Regression Tree (MART) based on boosting algorithm and decision stump provides the best accuracy.

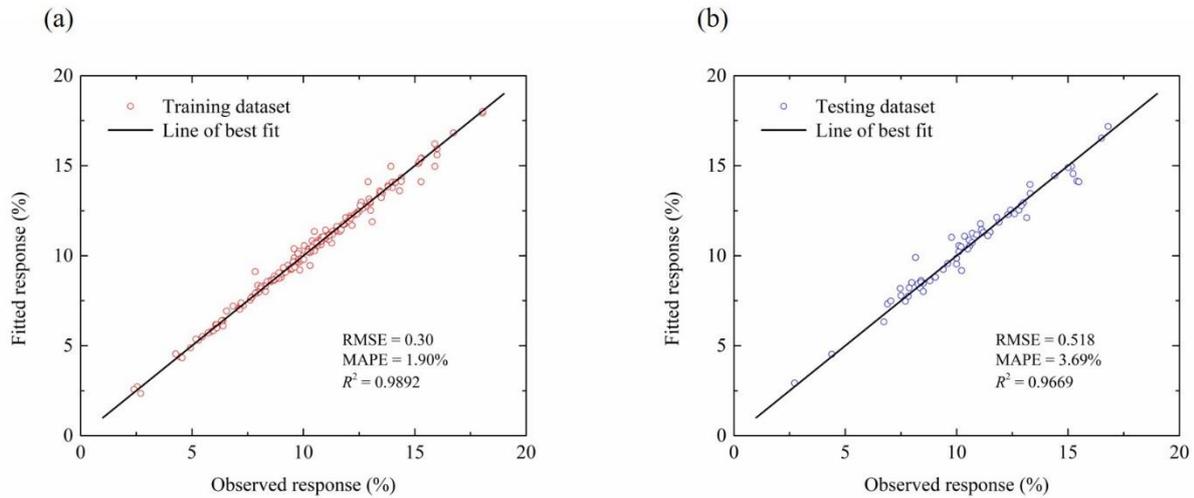

**Fig. 7**. Prediction performance of gradient boosting trees: (a) Training dataset; (b) Testing dataset

However, random forest produces very stable prediction accuracy, while gradient boosting may suffer from high variance in the random simulations. This is because random forest contains only two candidate hyperparameters that need to be tuned and the optimized parameters are quite similar across multiple runs. Random forest is also known to be a robust machine learning algorithm that performs well with its default settings [50]. Another advantage of random forest is that the *out-of-bag* (OOB) error based tuning strategy is much faster than *k*-Fold Cross-Validation. Because the average number of distinct observations in each bootstrapped sample is $0.632 \cdot n$ from Eq. (4), the OOB error will roughly behave like 2-Fold CV error [33]. Hence, unlike Gradient Boosting, Random Forests can be fitted in one sequence, with Cross-Validation being performed along the way.

One good thing with random forests is that the predictor importance can be estimated by permutation of *out-of-bag* (OOB) predictor observations for the ensemble of trees [33,51]. As briefly summarized in Fig. 8, the increase in OOB error as a result of randomly permutating the observations of *j*th predictor variable in the OOB sample is computed for each tree and then averaged over all trees to indicate the importance of variable *j* in the random forests. The influence of a variable in predicting the response increases with the value of this importance measure. For the concrete porosity data set studied in the present work, the variable importance plots are shown in Fig. 9. Curing days, binder content and *w/b* ratio are considered to be the most critical factors in predicting concrete porosity. These data-driven perspectives agree quite well with the published experimental works [13-15].

**Algorithm:** *Out-of-Bag* Predictor Importance Estimates by Permutation

**Train** random forests with $B$ learners on the training data with total $p$ predictor variables

**For** $b = 1$ **to** $B$ **do**:
- Identify the OOB observations and the subset of $J$ predictors ($J \leq p$)
- Estimate prediction accuracy on the OOB sample $oobError^b$
- **For** $j = 1$ **to** $J$ **do**:
  - Randomly permute the observations of $x_j$ in the OOB sample
  - Re-estimate prediction accuracy on the permuted $\widetilde{OOB}$ sample $\widetilde{oobError}_j^b$
  - Calculate the difference in accuracy $\Delta_j^b = \widetilde{oobError}_j^b - oobError^b$
- **End**

**End**

Compute mean $\overline{\Delta}_j$ and standard deviation $\sigma_{\Delta_j}$ for each predictor $x_j$ over $B$ learners

**Output:** $VI(x_j) = \overline{\Delta}_j / \sigma_{\Delta_j}$ for $j = 1, \ldots p$.

**Fig. 8**. *Out-of-bag* permuted predictor importance algorithm for random forests of regression trees

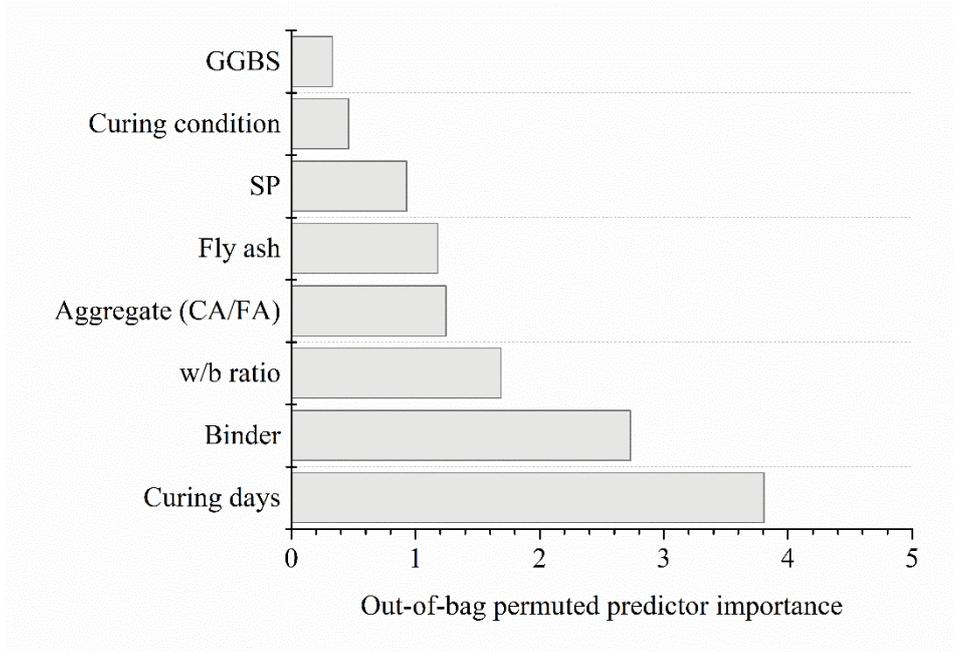

**Fig. 9**. Predictor importance for random forests

The partial dependent plot (PDP) shows the relationships between a predictor and the response of regression in the trained model. The partial dependence on the selected predictor is defined as the averaged prediction obtained by marginalizing out the effect of the other variables [33]. Fig. 10 displays the single-variable partial dependence plots on the four most relevant predictors including curing days, binder content, *w/b* ratio and aggregate (CA/FA ratio). The vertical scales of the plots are the same, and give a visual comparison of the relative importance of the different variables. Porosity is generally monotonic decreasing with increasing curing days, but gradually reaching steady after 150 curing days. Porosity has a non-monotonic partial dependence on binder content. The PDP shows a large change near binder = 405.6 ~ 408.6 kg/m$^3$, which indicates many node splits are based on this critical value of binder. The partial dependence of porosity on *w/b* ratio is monotonic increasing. The influence of aggregate composition on porosity is on a relatively small scale. It is important to note that there may exist some interaction between different predictors, which can't be illustrated through the single-variable PDPs.

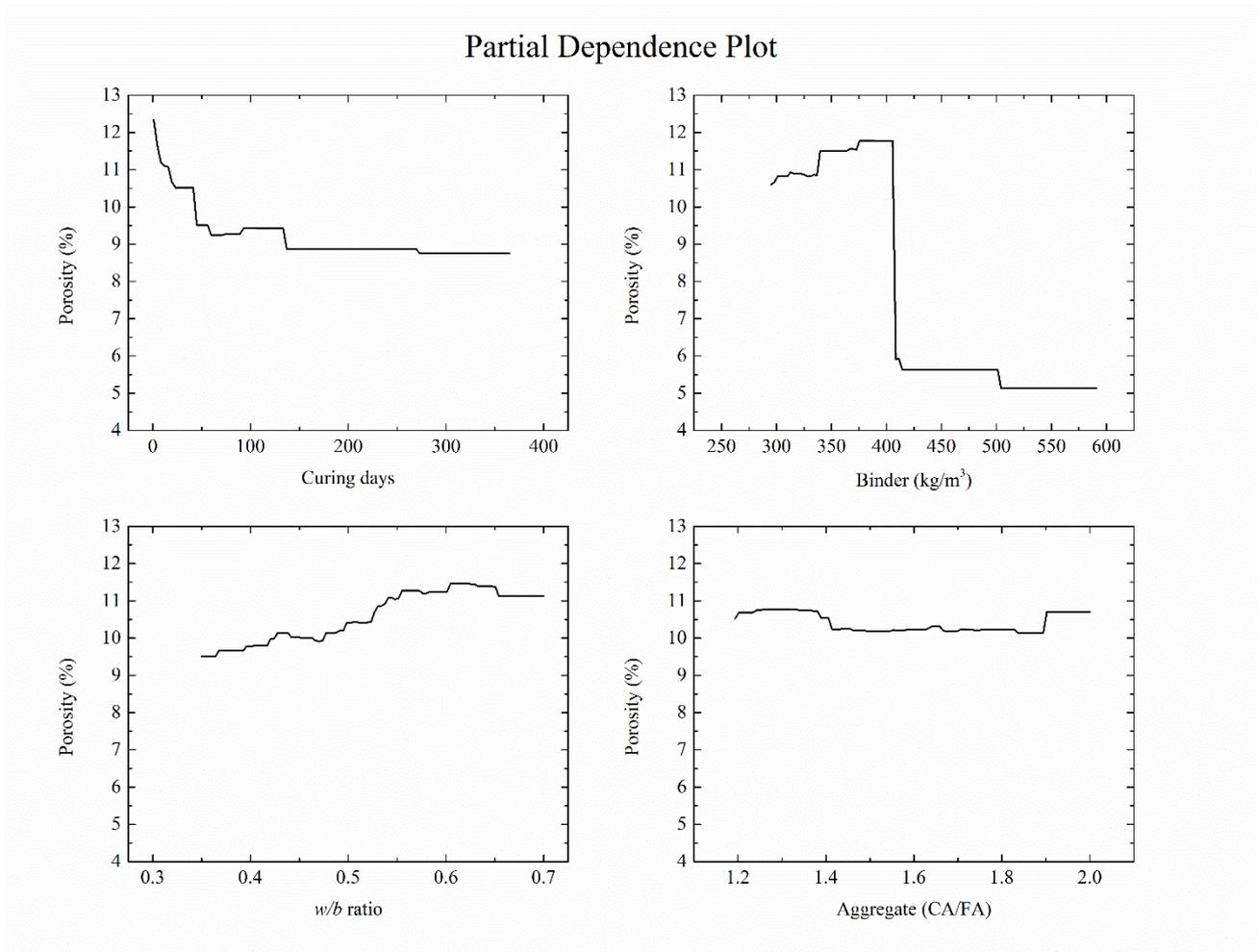

**Fig. 10**. Partial dependence plots on selected variables

## *4.2 Sensitivity analysis*

To further demonstrate the predicting capability of the proposed machine learning algorithm, sensitivity analysis has been performed on fly ash concrete and GGBS concrete. The predictions are made by employing the optimized gradient boosting tree. The artificial concrete mixture compositions are listed in Table 5. The varying parameters include fly ash or slag replacement portions and curing days.

As shown in Fig. 11 for fly ash concrete, the porosity increases with increasing fly ash portion in the artificial specimens at early ages. The addition of fly ash in concrete can reduce water content requirement for a given workability and provide refinement of pore structure. Due to the long-term nature of the pozzolanic reaction, the beneficial effect of fly ash become evident after 100 days of curing. The well-cured fly ash concrete has much lower porosity than OPC when other compositions are the same. This observation is also consistent with the long-term strength developments in fly ash concrete [57]. This phenomenon can also be demonstrated through the two-variable partial dependence plot of the fitted model on joint values of fly ash content and curing days, as shown in Fig. 12.

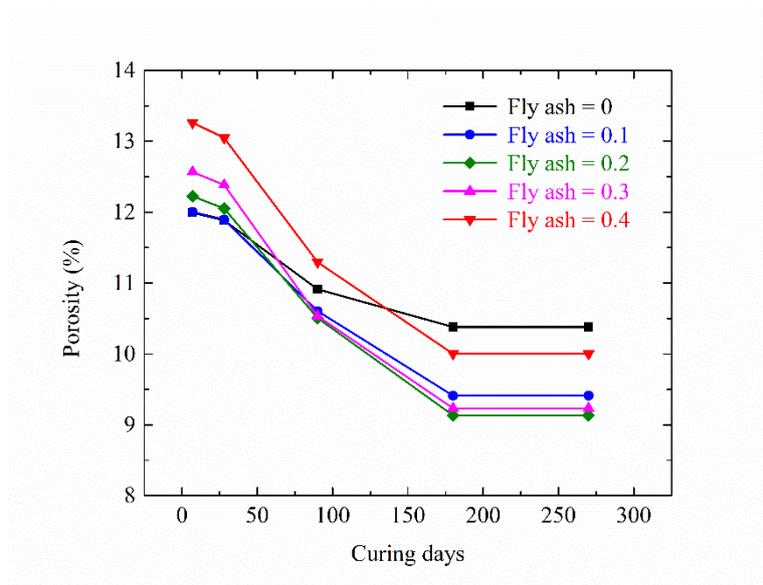

**Fig. 11**. Sensitivity test for fly ash concrete

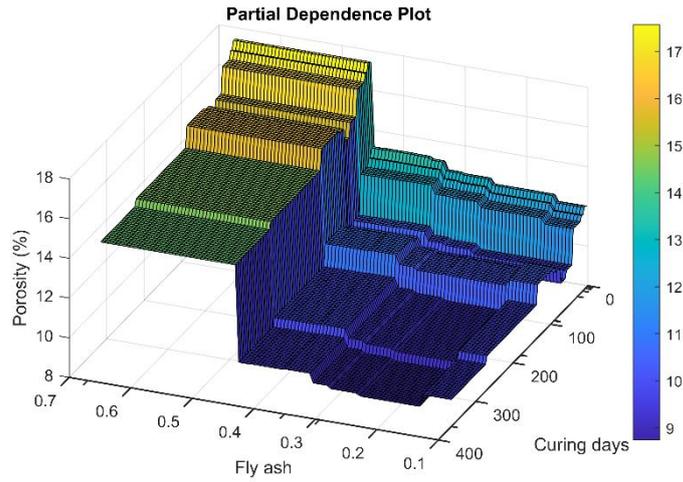

**Fig. 12**. Partial dependence plot on fly ash content and curing days for fly ash concrete

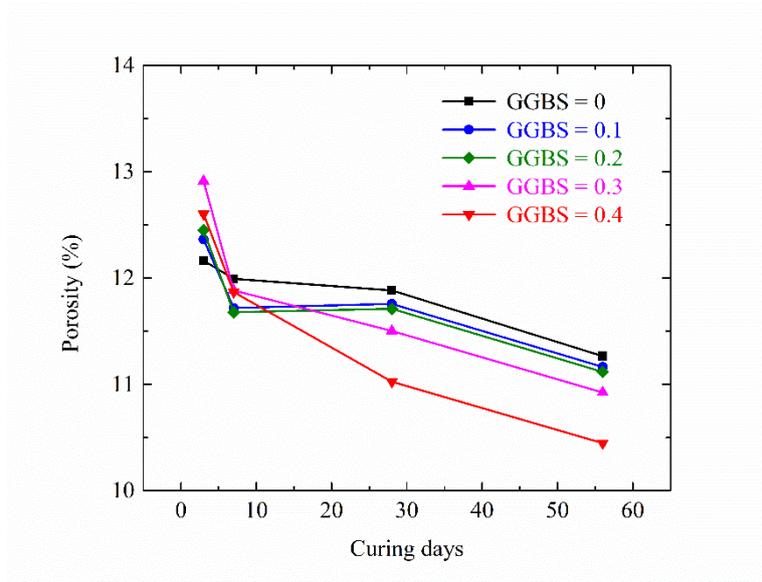

**Fig. 13**. Sensitivity test for GGBS concrete

The porosities of the artificial GGBS concrete specimens as a function of curing age are shown in Fig. 13. As the level of slag replacement increases, the early-age porosity increases. However, the long-term pore structure of concrete continues to refine as a result of the cement hydration process, which is contributed by the latent hydraulic reaction of GGBS. Given sufficient curing time, the concrete porosity generally decreases as the slag replacement ratio increases. Similar results have been observed by Choi *et al*. [58] in their experimental study of high-strength cement pastes incorporating high volume GGBS. The two-variable partial dependence of the regression model on joint values of GGBS content and curing days is shown in Fig. 14, which clearly demonstrate the strong interaction between these two variables.

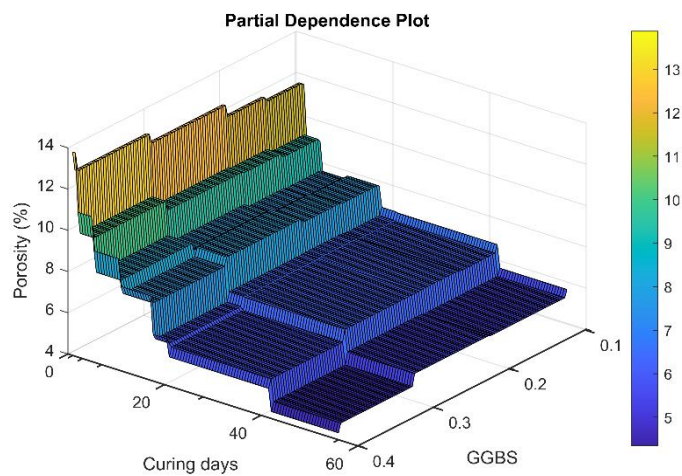

**Fig. 14**. Partial dependence plot on GGBS content and curing days for GGBS concrete

## *4.3 Discussion*

In this study, the 75/25 random partition of the porosity dataset is utilized for both Gradient Boosting Trees and Random Forests. The purpose is to compare the prediction performance of the two methods on the same testing subset, which is independent of the training data. On the training data set, Random Forests is optimized with *out-of-bag* error, while Gradient Boosting Trees is optimized with *k*-fold CV error. The evolution of training error (on the 75% training subset) and testing error (on the 25% testing subset) during the training process is plotted together with the *out-of-bag* error for Random Forests in Fig. 15a. The testing MSE settles down together with the OOB MSE as the number of trees grows, though the OOB error is relatively higher than the testing error. Similarly for Gradient Boosting Trees in Fig. 15b, the testing error and the *k*-fold CV error have converged with the latter relatively higher than the former. The simultaneous convergence of testing error and *out-of-bag* error or *k*-fold CV error clearly demonstrates that the proposed optimization algorithm could produce the best combination of hyperparameters.

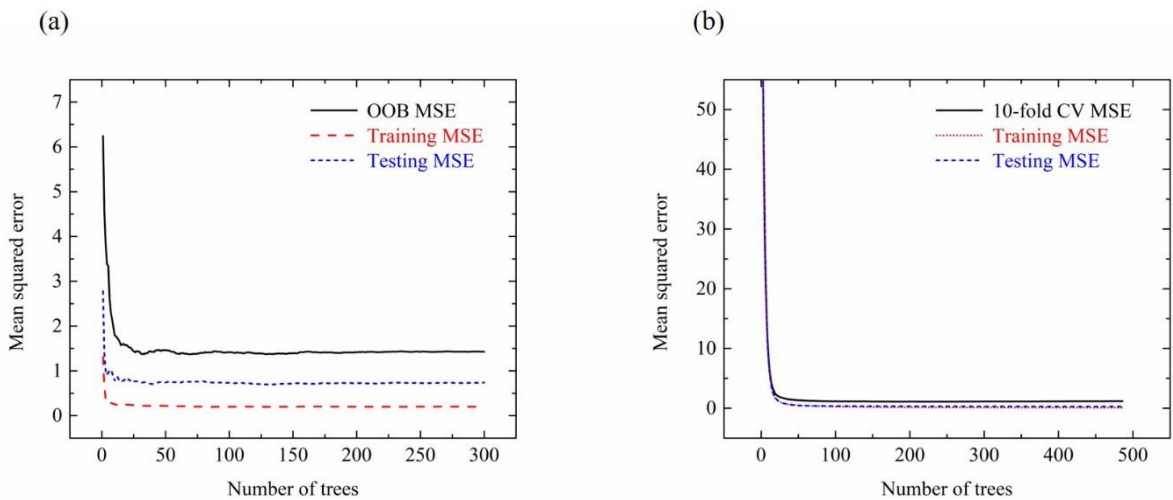

**Fig. 15**. The evolution of training error and testing error together with: (a) OOB error for Random Forests; (b) *k*-fold CV error for Gradient Boosting Trees

Unlike Gradient Boosting Trees, Random Forests doesn't necessarily require dividing the whole data set into 75/25 training/testing subsets because the 63/37 random partition is naturally embedded in the bootstrap sampling procedure [35]. The *out-of-bag* error can be conveniently used as a measure of prediction accuracy for the ensemble trees algorithm. However, there are some arguments among practitioners and researchers on whether *out-of-bag* error is an unbiased estimate of the generalized error for all datasets, though minimization of OOB MSE can always be used for tuning Random Forests to find the optimal hyperparameters [50,53,59]. For the concrete porosity training dataset, the *out-of-bag* prediction performance statistics based on the 100 simulations are summarized in Table 6. As compared with the testing performance metrics for Random Forests in Table 4, the *out-of-bag* error obtained from the training process is relatively higher than the prediction error on a new testing data set. Similar observation has been reported by Breiman [47], who has concluded that the average *out-of-bag* error is consistently higher than the average testing error for regression forests. Note that the sample instances as well as the number of samples are different for the two performance metrics. Hence, to ensure a fair comparison between Random Forests and Gradient Boosting Trees, testing error on the same testing data set is being used in this research.

In this research, the number of trees grown ($B$) is set to be not tunable, but sufficiently large, for Random Forests. Breiman [47] shows that the mean squared generalization error of random forest regression converges as the number of aggregated trees increases. This may suggest that the larger number of trees grown in the forests, the more accurate the ensemble prediction is. For the concrete porosity dataset, the performance gains in terms of *out-of-bag* error by increasing the number of trees from 50 to 1000 are illustrated in Fig. 16. The results are based on 100 simulations for each setting and each run involves hyperparameter optimization. The variance of OOB MSE significantly reduces as $B$ increases. The average value of OOB MSE over 100 simulations also shows a monotonously descending trend with the increasing $B$. However, there is only modest prediction improvement after growing the first 300 trees. Therefore, the choice of $B = 300$ would be sufficient. Moreover, in the case of a large dataset, the number of trees should be properly adjusted according to the computational cost of Random Forests optimization.

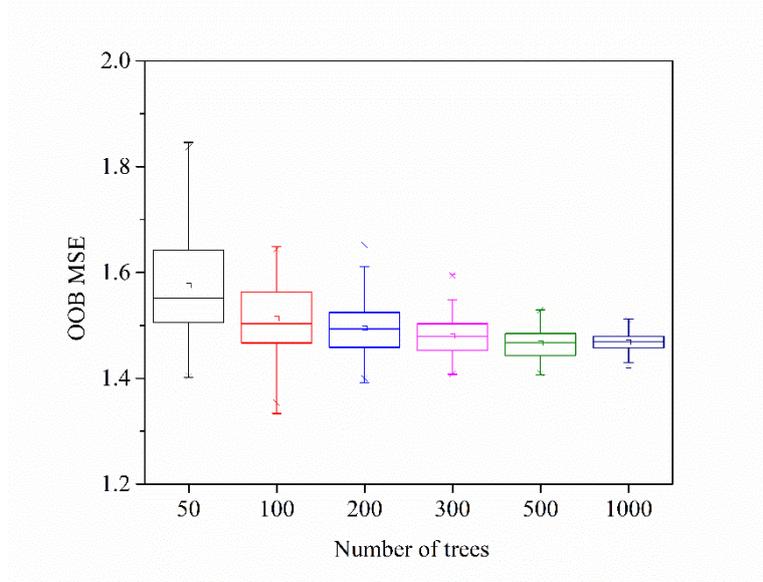

**Fig. 16**. The effect of number of trees on the *out-of-bag* error in random forest regression

# 5 Comparison with Chemo-Mechanical Model

Based on the chemo-mechanical modeling of Portland cement hydration and fly ash pozzolanic reactions, Papadakis [23] has proposed a theoretical model to predict the chemical and volumetric composition of fly ash concrete. The fly ash concrete porosity ($\varepsilon$) can be calculated as

$$\varepsilon = \varepsilon_{air} + W/\rho_w - \Delta\varepsilon_h - \Delta\varepsilon_p - \Delta\varepsilon_c \tag{9}$$

where $\varepsilon_{air}$ denotes the volume fraction of entrapped or entrained air in concrete, $W$ gives the water content in concrete mixture (kg/m$^3$), $\rho_w = 1000$ kg/m$^3$ is the density of water, and $\Delta\varepsilon_h$, $\Delta\varepsilon_p$, $\Delta\varepsilon_c$ represent the porosity reductions due to cement hydration, pozzolanic activity and carbonation, respectively.

By assuming the full hydration of cement and the complete pozzolanic reactions of fly ash, the final value of the porosity of a noncarbonated concrete can be predicted based on the physical and chemical properties of cement and fly ash. Denote as $f_{i,c}$ and $f_{i,p}$ ($i = C, S, A, F, \bar{S}$) the weight fractions of oxides CaO (C), SiO$_2$ (S), Al$_2$O$_3$ (A), Fe$_2$O$_3$ (F), SO$_3$ ($\bar{S}$) in cement and fly ash, respectively. The glassy phases constitute the reactive portion of fly ash, particularly in low-calcium fly ash. The active fractions of SiO$_2$ and Al$_2$O$_3$ in fly ash that contribute to the pozzolanic reactions are represented by $\gamma_S$ and $\gamma_A$ (by weight). The cement and fly ash content in concrete mixture are given by $C$ and $P$ (kg/m$^3$).

If the gypsum content is *higher* than that required for the full hydration of cement and the complete pozzolanic reaction of fly ash alumina, i.e.

$$f_{\bar{S},c} > 0.785 f_{A,c} - 0.501 f_{F,c} + (0.785 \gamma_A f_{A,p})(P/C) \tag{10}$$

The final value of the porosity of a noncarbonated concrete can be determined by the following equation [23]

$$\varepsilon = \varepsilon_{air} + W/\rho_w - \{0.249(f_{C,c} - 0.7 f_{\bar{S},c}) + 0.191 f_{S,c} + 1.118 f_{A,c} - 0.357 f_{F,c}\} \times (C/1000) \\ - (1.18 \gamma_A f_{A,p})(P/1000) \tag{11}$$

where the maximum fly ash content ($P_{max}$) that can participate in the pozzolanic reactions is specified as

$$P_{max} = \frac{\{1.321(f_{C,c} - 0.7 f_{\bar{S},c}) - 1.851 f_{S,c} - 2.182 f_{A,c} - 1.392 f_{F,c}\} \times C}{1.851 \gamma_S f_{S,p} + 2.182 \gamma_A f_{A,p}} \tag{12}$$

If the gypsum content is *lower* than that required for the full hydration of cement and the complete pozzolanic reaction of fly ash alumina, i.e.

$$f_{\bar{S},c} < 0.785 f_{A,c} - 0.501 f_{F,c} + (0.785 \gamma_A f_{A,p})(P/C) \tag{13}$$

The final value of the porosity of a noncarbonated concrete can be determined by the following equation [23]

$$\varepsilon = \varepsilon_{air} + W/\rho_w - (0.249 f_{C,c} - 0.1 f_{\bar{S},c} + 0.191 f_{S,c} + 1.059 f_{A,c} - 0.319 f_{F,c}) \times (C/1000)$$
$$- (1.121 \gamma_A f_{A,p})(P/1000) \tag{14}$$

where the maximum fly ash content ($P_{max}$) that can participate in the pozzolanic reactions is specified as

$$P_{max} = \frac{(1.321 f_{C,c} - 1.851 f_{S,c} - 2.907 f_{A,c} - 0.928 f_{F,c}) \times C}{1.851 \gamma_S f_{S,p} + 2.907 \gamma_A f_{A,p}} \tag{15}$$

Here, assumption has been made that the chemo-mechanical theory developed for pastes and mortars could also be applied to concrete.

The proposed case study is utilized to compare the conventional chemo-mechanical model with the machine learning method. To comply with the theoretical assumption of "complete" hydration and pozzolanic activities for Papadakis's model, we select 25 concrete specimens with at least 1-year hydration (365 curing days) from the assembled dataset. It is assumed that the entrained air in concrete is negligible ($\varepsilon_{air} = 0$) and that there is no carbonation. The chemical compositions of the cement and fly ash used in the original experimental work [40] are shown in Table 7. Since no information has been provided about the glass phase content of the fly ash constituents, the active fractions ($\gamma_S = \gamma_A = 0.82$) reported by Papadakis [23] are adopted in this study. In literature, the experimentally measured glassy phase compositions of fly ash vary a lot at both bulk scale and oxide level [60-62]. For example, Cho *et al*. [61] have reported amorphous phases in the range of 68.1% to 77.6% with an average of 73.0% for bulk fly ashes, $\gamma_S \in (66.6\%, 77.9\%)$ for SiO$_2$ and $\gamma_A \in (57.4\%, 77.2\%)$ for Al$_2$O$_3$.

The comparison between the analytical prediction from Papadakis model and the empirical prediction from Random Forest is shown in Fig. 17. There are 18 instances chosen from the training data set, for which *out-of-bag* predictions are used. This is to reduce the possible overfitting bias in the error estimation for the training data. It can be clearly observed that Random Forest easily outperforms the conventional chemo-mechanical model in terms of RMSE and MAPE. This suggests that the proposed data-driven approach could be applied for practical estimation of concrete porosity.

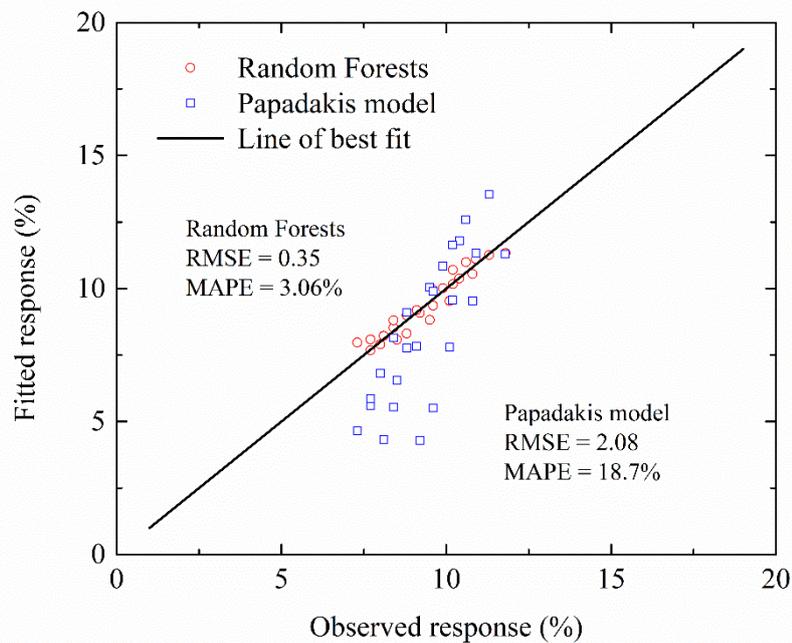

**Fig. 17**. Comparison of random forests with chemo-mechanical model

# 6 Conclusions

This paper applies ensemble trees to predict the porosity of high-performance concrete containing supplementary cementitious materials. A reliable database for concrete porosity, featuring 74 unique concrete mixtures, is assembled from published literature. Compositions of concrete are characterized by 8 features including *w/b* ratio, binder content, fly ash, GGBS, superplasticizer, coarse/fine aggregate ratio, curing condition and curing days. The full dataset is randomly divided into 75% training dataset and 25% testing dataset through stratified sampling.

The complexity (depth) of the individual trees grown in the ensemble can be regulated via limiting the maximum number of splits and/or the minimum size of terminal nodes. For boosted trees, the number of learning cycles can be controlled by the learning rate. In the case of random forest, training samples can be bootstrapped from the entire training data set and the predictors can be selected randomly at each split. The complexity level of boosted regression trees is tuned using $k$-Fold Cross-Validation, while the hyperparameters for random forest are optimized by minimizing *out-of-bag* (OOB) error. Bayesian optimization has been employed to search for the best combination of hyperparameters for regression tree ensembles.

Experimental tests show that ensemble trees can accurately predict the porosity of concrete from mixture compositions. Gradient boosting trees generally outperforms random forests in terms of prediction accuracy. Shallow trees are typically grown for gradient boosting, while full-depth trees work well for random forests. The OOB error based tuning strategy for random forest is found to be much faster than $k$-Fold Cross-Validation. The variable importance plot shows that curing days, binder content and *w/b* ratio are the most important predictors for concrete porosity. Sensitivity analysis further demonstrates the long-term beneficial effects of fly ash and GGBS in reducing concrete porosity.

Compared with conventional statistical regression or classical chemo-mechanical hydration model, the proposed ensemble learning algorithm is able to take into consideration the complex concrete compositions and achieve high prediction accuracy. Potential applications of this method may include optimizing concrete mixture compositions for performance-based concrete structure design and for reducing environmental impact of concrete. Future work should continue to build a reliable and balanced database to train the ensemble trees model. The prediction performance can be further improved by combining different machine learning algorithms.

# Data Availability

The training data and machine learning codes for this study are available upon request.

Table 1  Selected experimental data

| Reference | Mix ID | w/b | Binder (kg/m³) | Fly ash (%) | GGBS (%) | SP (%) | Aggregate (CA/FA) | Curing condition | Curing days | Porosity (%) | Training |
|---|---|---|---|---|---|---|---|---|---|---|---|
| Ahmad and Azad (2013) | 1 | 0.4 | 300 | 0 | 0 | 0 | 1.6 | air | 28 | 9.58 | True |
| | 2 | 0.4 | 350 | 0 | 0 | 0 | 1.6 | air | 28 | 11.08 | True |
| | 3 | 0.4 | 400 | 0 | 0 | 0 | 1.6 | air | 28 | 11.27 | True |
| | 16 | 0.6 | 300 | 0 | 0 | 0 | 1.8 | air | 28 | 11.07 | False |
| | 17 | 0.6 | 350 | 0 | 0 | 0 | 1.8 | air | 28 | 11.63 | True |
| | 18 | 0.6 | 400 | 0 | 0 | 0 | 1.8 | air | 28 | 12.64 | True |
| Shafiq et al. (2007) | UK0 | 0.55 | 325 | 0 | 0 | 0 | 1.5 | water | 3 | 11.12 | False |
| | UK30 | 0.49 | 325 | 30 | 0 | 0 | 1.5 | water | 3 | 10.80 | True |
| | UK40 | 0.48 | 325 | 40 | 0 | 0 | 1.5 | water | 3 | 10.79 | True |
| | MY0 | 0.56 | 325 | 0 | 0 | 0 | 1.5 | water | 180 | 9.77 | False |
| | MY30 | 0.525 | 325 | 30 | 0 | 0 | 1.5 | water | 180 | 8.38 | True |
| | MY40 | 0.515 | 325 | 40 | 0 | 0 | 1.5 | water | 180 | 8.27 | False |
| Van den Heede et al. (2010) | F0-1 | 0.5 | 350 | 0 | 0 | 0 | 1.32 | air | 28 | 15.40 | False |
| | F0-2 | 0.4 | 400 | 0 | 0 | 0.228 | 1.67 | air | 28 | 13.42 | True |
| | F35 | 0.4 | 400 | 35 | 0 | 0.228 | 1.67 | air | 28 | 15.23 | False |
| | F35 | 0.4 | 400 | 35 | 0 | 0.228 | 1.67 | air | 91 | 13.54 | True |
| | F50-4 | 0.4 | 400 | 50 | 0 | 0.3 | 1.67 | air | 91 | 12.90 | True |
| | F67 | 0.4 | 400 | 67 | 0 | 0.228 | 1.67 | air | 91 | 16.50 | False |
| Al-Amoudi et al. (1996) | – | 0.35 | 350 | 0 | 0 | 0 | 2 | air | 28 | 10.7 | False |
| | – | 0.35 | 350 | 20 | 0 | 0 | 2 | air | 28 | 11.3 | True |
| | – | 0.35 | 350 | 40 | 0 | 0 | 2 | air | 28 | 12.1 | True |
| | – | 0.55 | 350 | 0 | 0 | 0 | 2 | air | 365 | 11.8 | True |
| | – | 0.55 | 350 | 20 | 0 | 0 | 2 | air | 365 | 10.2 | False |
| | – | 0.55 | 350 | 40 | 0 | 0 | 2 | air | 365 | 11.3 | True |
| Younsi et al. (2011) | RefI | 0.60 | 301 | 0 | 0 | 0 | 1.27 | air | 28 | 16 | True |
| | FA30 | 0.53 | 341 | 30 | 0 | 0.51 | 1.27 | air | 28 | 15.9 | True |
| | RefI | 0.60 | 301 | 0 | 0 | 0 | 1.27 | water | 28 | 13.8 | True |
| | FA30 | 0.53 | 341 | 30 | 0 | 0.51 | 1.27 | water | 28 | 14.4 | True |
| Cheng et al. (2008) | WB35 | 0.35 | 591 | 0 | 0 | 0.7 | 1.7 | air | 7 | 6.40 | True |
| | WB35-20 | 0.35 | 591 | 0 | 20 | 0.7 | 1.7 | air | 7 | 6.39 | True |
| | WB35-40 | 0.35 | 591 | 0 | 40 | 0.7 | 1.7 | air | 7 | 6.84 | True |
| | WB70 | 0.7 | 296 | 0 | 0 | 0 | 1.2 | air | 56 | 7.66 | True |
| | WB70-20 | 0.7 | 296 | 0 | 20 | 0 | 1.2 | air | 56 | 7.13 | True |
| | WB70-40 | 0.7 | 295 | 0 | 40 | 0 | 1.2 | air | 56 | 5.72 | True |

Table 2  Statistical summary of continuous variables in the whole dataset

| Attributes | Minimum | Maximum | Mean | Standard deviation |
|---|---|---|---|---|
| *w/b* ratio | 0.35 | 0.7 | 0.48 | 0.10 |
| Binder content (kg/m$^3$) | 295 | 591 | 370 | 74 |
| Fly ash content (%) | 0 | 67 | 15 | 17.6 |
| GGBS content (%) | 0 | 40 | 4.4 | 10.6 |
| Superplasticizer (%) | 0 | 1.58 | 0.1 | 0.24 |
| Aggregate (CA/FA) | 1.2 | 2 | 1.7 | 0.3 |
| Curing days | 1 | 365 | 89 | 109 |
| Porosity (%) | 2.39 | 18.05 | 10.36 | 2.88 |

Table 3  Hyperparameters for regression trees ensemble

| Hyperparameters | Random Forest | Gradient Boosting |
|---|---|---|
| Number of trees | 300 | [10, 500] |
| Maximum number of splits | 179 | [1, 20] |
| Minimum size of terminal nodes | [1, 20] | [1, 90] |
| Number of features to select for each split | [1, 8] | 8 |
| Learning rate | N/A | [0.001, 1] |

Table 4  Summary of prediction performance statistics over 100 simulations

| Statistics | Random Forest | | | Gradient Boosting | | |
|---|---|---|---|---|---|---|
| | RMSE | MAPE | $R^2$ | RMSE | MAPE | $R^2$ |
| Minimum | 0.86 | 5.50% | 0.894 | 0.52 | 3.36% | 0.872 |
| Maximum | 0.93 | 6.10% | 0.909 | 1.02 | 7.12% | 0.967 |
| Mean | 0.89 | 5.77% | 0.901 | 0.68 | 4.66% | 0.942 |

Table 5  Concrete compositions for sensitivity test

| Attributes | Fly ash concrete | GGBS concrete |
|---|---|---|
| *w/b* ratio | 0.4 | 0.4 |
| Binder content (kg/m$^3$) | 400 | 400 |
| Fly ash content (%) | 0, 10, 20, 30, 40 | 0 |
| GGBS content (%) | 0 | 0, 10, 20, 30, 40 |
| Superplasticizer (%) | 0 | 0 |
| Aggregate (CA/FA) | 2 | 2 |
| Curing days | 7, 28, 90, 180, 270 | 3, 7, 28, 56 |
| Curing condition | air | air |

Table 6  Random Forests *out-of-bag* prediction performance statistics over 100 simulations

|  | RMSE | MAPE | $R^2$ |
|---|---|---|---|
| Minimum | 1.19 | 7.66% | 0.807 |
| Maximum | 1.26 | 8.34% | 0.830 |
| Mean | 1.22 | 7.90% | 0.822 |

Table 7  Chemical characteristics of cement and low-calcium fly ash [40]

| wt. % | $SiO_2$ | $Al_2O_3$ | $Fe_2O_3$ | CaO | $SO_3$ | MgO | $Na_2O$ | $K_2O$ | L.O.I. |
|---|---|---|---|---|---|---|---|---|---|
| Cement | 22.3 | 3.6 | 3.6 | 64.6 | 1.9 | 2.1 | 0.1 | 0.2 | 1.2 |
| Fly ash | 60.5 | 23.0 | 7.5 | 2.1 | 0.3 | 1.0 | - | - | 1.4 |